\documentclass{article}

\usepackage{arxiv}

\usepackage{latexsym}
\usepackage{amssymb}
\usepackage{amsmath}
\usepackage{amsthm}
\usepackage{subcaption}
\usepackage{booktabs} \newcommand{\ra}[1]{\renewcommand{\arraystretch}{#1}}
\usepackage{enumitem}
\usepackage{graphicx}
\usepackage{color}

\usepackage[utf8]{inputenc} 
\usepackage[T1]{fontenc}    
\usepackage{hyperref}       
\usepackage{url}            
\usepackage{amsfonts}       
\usepackage{nicefrac}       
\usepackage{microtype}      
\usepackage{cleveref}       
\usepackage{natbib}
\usepackage{doi}

\usepackage[dvipsnames]{xcolor} 
\usepackage{inconsolata} 
\usepackage{colortbl}

\definecolor{graygreen}{rgb}{0.8, 1, 0.8}

\title{Evaluating the Impact of Advanced LLM Techniques on AI-Lecture Tutors for a Robotics Course}

\newif\ifuniqueAffiliation

\ifuniqueAffiliation 
\author{ \href{https://orcid.org/0000-0000-0000-0000}{\includegraphics[scale=0.06]{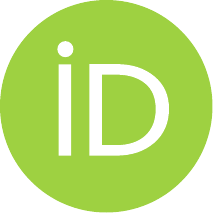}\hspace{1mm}David S.~Hippocampus}\thanks{Use footnote for providing further
		information about author (webpage, alternative
		address)---\emph{not} for acknowledging funding agencies.} \\
	Department of Computer Science\\
	Cranberry-Lemon University\\
	Pittsburgh, PA 15213 \\
	\texttt{hippo@cs.cranberry-lemon.edu} \\
	\And
	\href{https://orcid.org/0000-0000-0000-0000}{\includegraphics[scale=0.06]{orcid.pdf}\hspace{1mm}Elias D.~Striatum} \\
	Department of Electrical Engineering\\
	Mount-Sheikh University\\
	Santa Narimana, Levand \\
	\texttt{stariate@ee.mount-sheikh.edu} \\
}
\else
\usepackage{authblk}

\newbox{\orcid}\sbox{\orcid}{\includegraphics[scale=0.06]{orcid.pdf}}

\author[1]{%
    \href{https://orcid.org/0000-0002-8468-2808}{\usebox{\orcid}\hspace{1mm}Sebastian Kahl}%
}
\author[2]{%
    Felix Löffler%
}
\author[2]{%
    Martin Maciol%
}
\author[2]{%
    Fabian Ridder%
}
\author[2]{%
    Marius Schmitz%
}
\author[2]{%
    Jennifer Spanagel%
}
\author[2]{%
    Jens Wienkamp%
}
\author[1]{%
    Christopher Burgahn%
}
\author[2]{%
    \href{https://orcid.org/0000-0002-0849-483X}{\usebox{\orcid}\hspace{1mm}Malte Schilling\thanks{\texttt{malte.schilling@uni-muenster.de}}}%
}
\affil[1]{fronyx GmbH, Rüttenscheider Str. 120, 45131 Essen}
\affil[2]{Autonomous Intelligent Systems Group, University of Münster}
\fi

\renewcommand{\shorttitle}{Advanced LLM Techniques for AI-Lecture Tutors}

\hypersetup{
pdftitle={A template for the arxiv style},
pdfsubject={q-bio.NC, q-bio.QM},
pdfauthor={David S.~Hippocampus, Elias D.~Striatum},
pdfkeywords={First keyword, Second keyword, More},
}

\begin{document}
\maketitle

\begin{abstract}
This study evaluates the performance of Large Language Models (LLMs) as an Artificial Intelligence-based tutor for a university course. In particular, different advanced techniques are utilized, such as prompt engineering, Retrieval-Augmented-Generation (RAG), and fine-tuning. We assessed the different models and applied techniques using common similarity metrics like BLEU-4, ROUGE, and BERTScore, complemented by a small human evaluation of helpfulness and trustworthiness. Our findings indicate that RAG combined with prompt engineering significantly enhances model responses and produces better factual answers. In the context of education, RAG appears as an ideal technique as it is based on enriching the input of the model with additional information and material which usually is already present for a university course. Fine-tuning, on the other hand, can produce quite small, still strong expert models, but poses the danger of overfitting. Our study further asks how we measure performance of LLMs and how well current measurements represent correctness or relevance? We find high correlation on similarity metrics and a bias of most of these metrics towards shorter responses. Overall, our research points to both the potential and challenges of integrating LLMs in educational settings, suggesting a need for balanced training approaches and advanced evaluation frameworks.
\end{abstract}


\section{Introduction}
Large Language Models have become popular with the introduction of ChatGPT. In particular, the novel dialogue capabilities of these Assistant Models \citep{ouyang2022training} has spurned an interest as these provide a flexibility that allows to apply LLMs in various sectors, including education. These Large Language Models (LLMs) are Transformer-based neural networks \citep{vaswani2017attention} that were trained on massive text corpora and that exhibit impressive abilities to comprehend, process, and generate human-like language. Originally, Large Language Models---now often called Foundation Models---were trained with the goal to predict the following text token, taking into account long and diverse relationships of the given previous input. The models learnt from this to perform well on a large number of different tasks across a variety of topics. Importantly, early work on GPT-2 already demonstrated the flexibility of these models as these models can be explicitly directed to execute a specified language task simply through prompting these models accordingly \citep{radford2019language}. Early-on, these models had difficulties staying on track for a given task and often their responses lingered or drifted away from a topic. As a consequence, the earliest application were in specific and narrowly defined tasks as translation. In such a task the model could continuously be forced to follow the original input sequence. The work by \cite{ouyang2022training} changed this, as they further trained models to better imitate human behavior which lead to the emergence of Assistant Models as ChatGPT. These demonstrated for the first time the ability to provide a--to a certain degree--consistent partner for dialogue which can be engaged in a conversation. Modern models are now able to understand and generate language with remarkable fluency and coherence. 
The success of such models has now spread and there is a growing number of models similar to ChatGPT that are further improving at a rapid pace (e.g., Meta's LLaMA-2 as an open-source model \citep{touvron2023llama2}; for an overview on the current landscape of LLMs see \citep{naveed2024comprehensive}). 
LLMs present now valuable opportunities to design and develop innovative assistant systems.

In the area of education, the early debate on the impact of powerful LLMs has sadly mostly focused on negative consequences, e.g., how this might allow students to cheat as these models are capable of generating written course work---as currently used to test learning capabilities---or even parts or a full thesis work. 
But LLMs can also be beneficial in education and teaching: LLMs as an assistant can be utilized to provide individualized teaching for every student. In particular, the possibility to prompt them to answer in a certain role---e.g., as a primary or high school teacher---and addressing a specific audience---e.g., a fourth grader or a grad student---offers novel possibilities for better adapted teaching that takes the need of the individual learner into account \citep{extance2023chatgpt}. This has lead to the idea of introducing LLMs as tutors for courses or lectures. There are already quite a number of different nice examples. One particular area is the area of coding assistants (e.g., Starcoder by Hugginface \citep{starcoder2023}; see as well \citep{mehta2023can} and for a systematic overview \citep{liu2023large}). Other examples include assistants in language learning. Khan Academy's assistant Khanmigo \citep{khanmigo2023} is an early example in this area as they started to develop this in cooperation with OpenAI before the official release of ChatGPT. In general, Khan Academy has embraced LLMs as powerful tools for education and their founder has detailed a positive outlook how this not only can help to provide better teaching---in the sense of more individualized tutoring that supports learning---, but as well more engaging learning experiences \citep{khan2024bravenewwords}. Following such a route, we are seeing an increasing number of LLM-based systems acting as tutors in university courses (as one example, see  \citep{berkeley2023oatutor}; for a broad review see \citep{garcia2024review}). 

A significant challenge in deploying LLMs are hallucinations. Hallucinations in LLMs refer to cases in which the model generates plausible-sounding, but incorrect or nonsensical responses \citep{maynez-etal-2020-faithfulness}. It is assumed that this occurs because the model relies solely on patterns in textual data it was trained on without any understanding of factual correctness \citep{huang2023survey}. Detecting hallucinations has proven difficult, but different techniques and metrics have been proposed (for a comprehensive review see \citep{huang2023survey} which outlines principles, a taxonomy, and open questions regarding hallucinations in LLMs, while as well providing insights on developing effective detection and mitigation strategies). The risk of hallucinations is particularly concerning for an educational setting and has to be addressed in such an area that demands factual correctness. 

One major focus in utilizing LLMs in education as a tutoring system is, therefore, in how to prevent hallucinations and restrict answers to the specific course \citep{dong2023aitutor,extance2023chatgpt}. Our work is directed in this direction, as we aim for an LLM as a tutor for learning and in teaching of a University course on robotics. This LLM-powered tutor was envisioned as a supplementary resource, providing tailored assistance and facilitating personalized learning paths for participants. The explicit goal of the presented work was to develop a tutor-like system that is utilizing the flexibility of Assistant Models. By this, we want to address two main research questions: First, LLMs can show the tendency to provide wrong or hallucinated responses. We wanted to analyze how different---currently widely proposed---advanced techniques for LLMs can improve the responses of LLMs in a specific knowledge domain? Do such extended LLMs hallucinate less and provide more factual answers? Secondly, evaluation of the performance of LLMs has shown to be difficult and lead to inconsistent results. Therefore, we want to address how can we measure performance of LLMs and how good do currently applied measurements represent factuality? In particular, we are interested in how do different metrics relate to each other or overlap? Furthermore, can we identify biases that affect these measurements?




In this paper, we want to summarize our experience with current State-of-the-Art techniques in the area of LLMs and demonstrate how these differentially impact the behavior of the lecture tutoring system. 
Therefore, the paper covers current methods all applied in the area of a lecture tutor for an introductory robotics course. First, simply using an of-the-shelf LLM that is purely put to the task through prompt engineering. Secondly, enriching the input by course material through Retrieval-Augmented-Generation in which content from slides and the annotations of the lecture videos were also provided to the model. Third, an LLM. was fine-tuned for the specific task on a generated set of question-answer pairs. 
The article is organized in the following way:
We first provide a detailed overview of our AI lecture tutor, explaining the design requirements, the different applied LLM extension techniques and how they are realized, the selected LLMs, as well as details on dataset and preparation.%
Then we present our results on an introduced test data set specific to the robotics course. One crucial aspect of our work is the evaluation of LLMs which still appears in its infancy. We describe our multi-faceted approach, combining automated benchmarks to quantify performance and a human-centered evaluation using questionnaires to garner qualitative insights into student interaction with the AI Lecture Tutor. Finally, we conclude with a critical reflection of our findings and highlighting potential avenues for future research in the domain of AI-powered educational technologies.


\section{Methods: Large Language Model as an AI Tutor}
The proposed AI tutor will be integrated into the learning platform of the lecture and is designed as a chatbot that allows students to communicate via written messages. Regarding its function, the tutor system should consistently respond in a polite and helpful way, as a human tutor would do. It needs to be able to understand questions given in natural language and to generate answers that are both syntactically and semantically accurate. Most importantly, the answers should be consistent with the lecture which means, answers are, first, correct and the system should not tell falsehoods. Easy and general questions should be answerable based on the pre-trained model.
Secondly, as the tutor should support the student in learning the concepts, the answers should relate to the content of the lecture as well as the way it is taught, e.g., picking up examples or providing references to content as the slides. Therefore, slides of the lecture as well as the transcripts of the recordings should be taken into account. 

These requirements guide the realization of the tutor system and in the following, we will introduce different advanced techniques for LLMs that will be applied in the tutoring system. We are interested in comparing how well an LLM can work as a lecture tutor. This requires an experimental setup in which different variations of the tutor are questioned in a systematic way and evaluated. In the following, we will introduce the necessary parts for such an evaluation and comparison. First, we introduce the different advanced techniques applied to the LLM, i.e., prompt engineering, Retrieval-Augmented-Generation, and fine-tuning. Secondly, we will briefly describe the preprocessing of the lecture material and, third, how we derived a test data set for evaluation and a data set for training. Fourth, we will briefly provide details on the used models and the fine-tuning process. Last, we introduce different evaluation metrics for the comparison. 

\subsection{Advanced LLM Techniques}
Applying an off-the-shelf dialogue LLM as a lecture tutor can serve as a baseline. But it has to be assumed that it provides generic answers that---while hopefully will be correct---are not tailored to the specific course. There is now already an established set of different approaches for improving answers of an LLM. We will briefly describe these different approaches that will later be compared in our evaluation.


\subsubsection*{Guiding the Model: Prompt Engineering} 
A simple improvement can be achieved by enriching the prompt given by the user and tailoring the prompt towards the particular task as well as the given setting. 
Enriching the prompt can guide the model to generate more relevant and accurate answers \citep{radford2019language}. 
To improve a prompt, it is first necessary to understand what is required by the model acting as a tutor: The tutor is expected to answer as a human tutor would, which means it only tells the truth and cites it sources. Therefore, these aspects need to be incorporated into the prompt. Achieving the desired outcome through prompt engineering is often an iterative process in which a given prompt is tried, evaluated and iteratively improved \citep{jones2023teaching}. We achieved good results with initially starting from a very simple, basic system message and improve this stepwise to include more details: The AI Tutor was prompted that it is a lecture assistant, an expert in the field of robotics and should only answer according to the context of the lecture. 
\footnote{"You are now a lecture assistant that helps students to understand the lecture. 
You are an expert about the lecture topic autonomic systems and mobile robots. 
You will always bring the answer in the context of the lecture."
\url{https://zivgitlab.uni-muenster.de/schillma/lecturellm/-/blob/main/prompts.py?ref_type=heads\%23L74}}
    
As the initial prompt didn't suffice to turn the model into a tutor-like personality, additional traits as helpful, respectful, and honest were added. Furthermore, the model was explicitly instructed to exclude any form of racist, harmful, toxic, dangerous, illegal, or unethical content. 

One particular problematic area when using a standard LLM was the tendency to always try to generate some output, even though a question was not well understood or the model simply didn't know the necessary facts for providing an answer. When providing additional information using RAG (see next section), as an additional instruction for the prompt it was added to restrict answers to the provided material.
\footnote{
"You are a helpful, respectful, and honest teaching assistant for lecture material that will be provided to you.
Your current task is to answer student queries. Always answer as helpfully as possible, while being safe.
Your answers should not include any harmful, unethical, racist, sexist, toxic, dangerous, or illegal content.
Please ensure that your responses are socially unbiased and positive in nature. If a question does not make sense,
or is not factually coherent, explain why instead of answering something incorrect.
If you don't know the answer to a question, please do not share false information. You only have access to the lecture material provided to you.
Lecture material includes slides, images, and transcripts of videos.
The only thing you can do is write text, nothing else! Never, ever repeat yourself."
\url{https://zivgitlab.uni-muenster.de/schillma/lecturellm/-/blob/main/prompts.py?ref_type=heads\%23L80}
}

As one further aspect, prompt engineering can be used to advise the model explicitly to break down a task or question and derive an answer in a step-wise manner. While this could be driven by an external loop calling a LLM multiple times and initially asking the model to divide the problem which means to generate questions that need to be answered beforehand or look-up necessary definitions \citep{wei2022chain}. 

\begin{figure}[t!]
\centering
\includegraphics[scale=0.5]{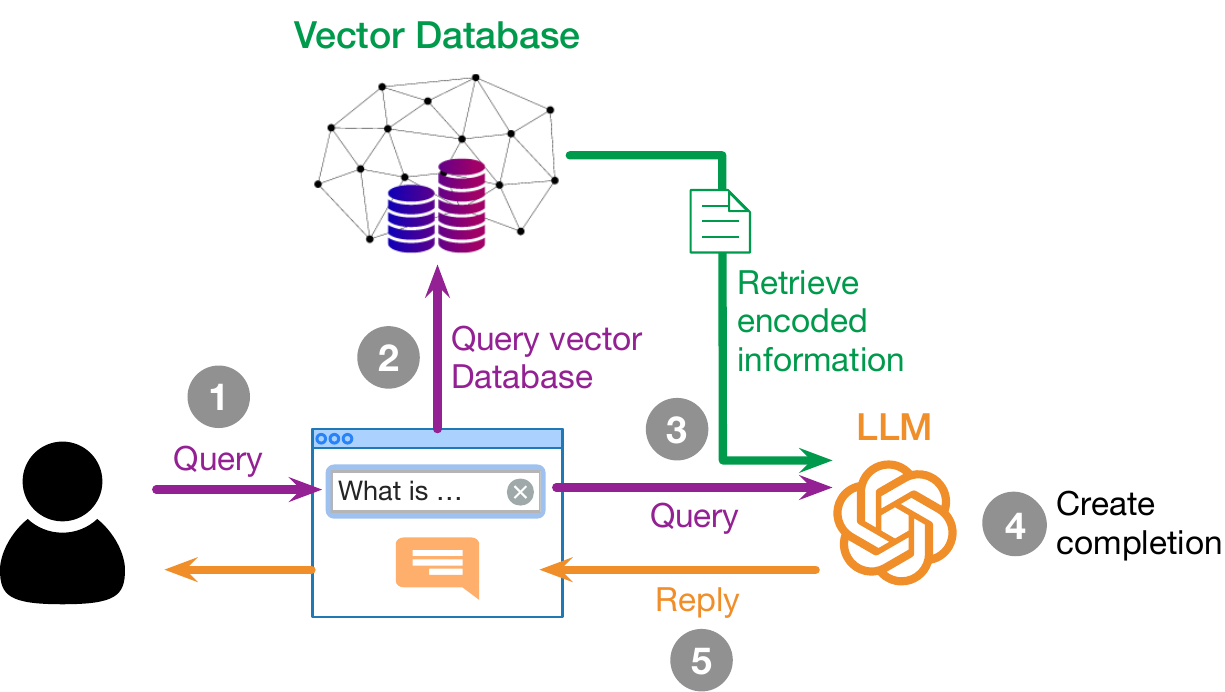}%
\caption{Overview of process for integrating RAG into the query to an LLM: The user asks a question (1) through an interface. For the embedded question additional information is retrieved from a vector database (2). The query to the LLM (3) integrates the original posed question and retrieved information into a---usually engineered---specific prompt. The LLM generates a completion as a response (4). Often, the reply has to be unwraped and can be passed to the user (5).}
\label{fig_2_rag}
\end{figure}

\subsubsection*{Enhancing the Input: Retrieval-Augmented Generation} 
One common extension for LLMs is to enrich the input to the system using external background data that relates to the given input and could potentially help in generating a response which is called Retrieval-Augmented Generation (RAG) \citep{lewis2021retrievalaugmented}.
The main idea of RAG, visualized in Fig. \ref{fig_2_rag}, is to retrieve such related information from a provided database. The retrieved data is added to the prompt as additional context. \cite{lewis2021retrievalaugmented} have shown that a model, that is supported with RAG, can outperform state-of-the-art approaches in open-domain Question Answering. It was also shown that such a model can approach the efficacy of systems with domain-specific architectures in Fact Verification.

RAG appears as an ideal extension for a tutoring system as it 
enriches responses with additional knowledge about the query, ensuring that answers are not only accurate but also relevant to the specific content covered in the lecture. 
This approach can also prevent the inclusion of overly detailed information not discussed in the lecture. Additionally, RAG's advancements in fact verification enable the AI tutor to identify and rectify inaccuracies presented by users during chat conversations. Lastly, RAG facilitates the integration of references into answers, for example, in our case, references to specific slides were included in the underlying database. This enables users to verify information, which should strengthen their trust in the model.

As a method, Retrieval-Augmented Generation can be divided into two primary stages (for more details, see \citep{zhao2024retrieval}): indexing and querying. The indexing stage involves the efficient storage of supplementary information. As an initial step, the data has to be segmented into manageable chunks which facilitates the search for specific content within the data. To enable semantic searches, a vector embedding is computed for each of these chunks. The database must store individual chunks and embeddings. Importantly, we want to be able to query the database semantically, which means it has to allow for a similarity search in the vector embedding space. 
Such a system is referred to as vector storage. \footnote{Vector storages: \url{https://python.langchain.com/docs/modules/data_connection/vectorstores } (Accessed: 28.02.2024)}


\begin{figure}[tbh]
\centering
\includegraphics[scale=0.5]{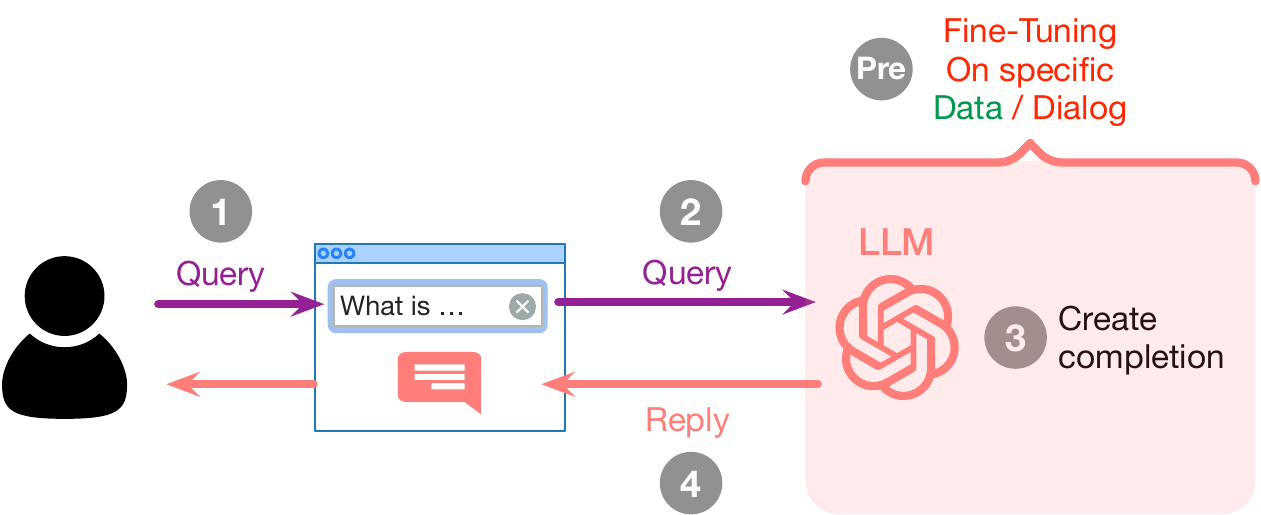}%
\caption{Overview of the query process when using a fine-tuned model (the access process is identically to directly querying a given LLM, but can of course be extended to as well include RAG): The user asks a question (1) through an interface which is given to the LLM (2). The LLM generates a completion as a response (3) which is passed to the user (4). The important difference is that task specific knowledge or interaction specific patterns are entrained into the model beforehand, i.e. during the fine-tuning stage (shown on the right in red).
}
\label{fig_3_finetune}
\end{figure}

\subsubsection*{Training the LLM: Fine-Tuning on Example Data}
Large Language Models operate on two sources of knowledge: As a parametric form, information is entrained into the distributed neural network structure of the model during training. Relatively huge amounts of text are used during training that gradually shape the responses of the model. As a second form, LLMs integrate contextual knowledge given as an input which drives the generation of the response. RAG explicitly provides additional contextual knowledge and in prompt engineering contextual input (the prompt) is used to steer the generation process (see above). A different, but usually more involved approach, is to address the knowledge stored in a distributed fashion in the neural networks of the LLM. This can be achieved using further training and fine-tuning the model \citep{devlin-etal-2019-bert}.
Fine-tuning a large language model involves adjusting the model's parameters and training it on a specific dataset to adapt it to a particular topic, in our case, the lecture on robotics. In fine-tuning, the neural network-based model is further trained in a supervised fashion on a specific data set. The goal is to to adapt the LLM towards a specific task and the model should become an expert in this task. Fine-tuning has shown to work well with relatively small data sets (of around a couple of hundreds to thousands example interactions), but when training on such a small data set, there is a danger of overfitting to these particular questions. The model might end up only providing memorized answers from this data set. In general, fine-tuning as a process can improve the model's performance and accuracy for a targeted use case by by combining the knowledge of the base model with specific information from a new dataset. 


\subsection{Preprocessing of Lecture Material}
Integration of lecture material into the query process using RAG or deriving training data from it, requires significant preprocessing of the available information. First, the information has to be extracted, in our case from slides which consist of written text that could be directly extracted from the original files (in our case based on PDFs of the lecture slides). In addition, graphical information and images from the slides were extracted and transferred to textual descriptions using multimodal Large Language Models. The visual information was fed into LLMs (GPT-4) that were tasked with briefly describing the given figure. Secondly, the original audio was taken from lecture videos in which the specific slides were explained and discussed in more detail. This requires transcribing the audio (realized using OpenAI's Whisper framework) and aligning the oral explanations with the content presented visually on the slides. The audio showed important providing complementing information and more detailed explanations. As a result of this process, we end up with a vast amount of data for eleven lectures (each one $90$ minutes long, roughly around $45$ slides per lecture) that is available to the system. During operation, the AI tutor should focus solely on relevant parts from the lecture and, therefore, needs an understanding of which different parts of the lecture and transcripts are necessary to answer the question. As an additional requirement, the AI tutor should provide information on its sources, e.g., pointing out which slides were used for a given answer (for example: "@10-slam-deck Slide 11"). This allows the user to check slides and verify the given answer.

Overall, the complete process should be executed quickly in general and efficiently to avoid compromising the user experience. For future application in real courses, factors regarding its scalability should be considered as well (both with respect to running an LLM as well as the necessary infrastructure for serving up background information from a database).

\subsection{Test and Fine-Tuning Dataset}
Evaluation of the different LLM techniques and models was done on the test data set. This data set consisted of chats (overall $478$) which each included a dialogue consisting of the question and a possible answer. Questions in the test set (and similarly in the training data set described below) were targeted for the specific robotics course and based on the material covered. In case of the test set, two lectures (on the topic of SLAM and navigation in robotics, lecture nine and ten) were used. Three students worked through the material of these lectures and generated by hand questions for the covered material which was then used again by the human writers to formulate an answer. Overall, for the test set $478$ such chats were written. 

For fine-tuning a training data set for the robotics course was generated. It explicitly excluded the topics covered in the test set for which there where humanly written chats. But for generating the test set, a LLM was used which was advised through a prompt to generate question-answer pairs for specific slides (the prompt had quite detailed instructions on how to generate questions based on the content of a slide which could then be used in addition to generate a fitting answer that included a reference to that particular slide). In addition, an example question-answer pair written by a human was given in a form of one-shot learning. The generated chats were afterwards read and checked by humans and were found to be generally of sufficient quality. This produced overall $2791$ chats for training that did not overlap with the test data set. The LLaMA-2 model was trained on this specific dataset of prototypical interaction. 

\subsection{Models and Fine-Tuning}
There are many LLMs openly available which are accessible through specific APIs and even more so through standard libraries and frameworks like Huggingface or their APIs. For our tutor system, we focussed, first, on GPT-3.5 which is accessible through OpenAI and for which we used the OpenAI API for access (model can be queried through simple python function calls). We chose this particular model as it is widely used and can be considered a good, common benchmark. It provides high quality output. Prompt engineering and RAG can be applied easily. 

For RAG, the AI Tutor application uses a PostgreSQL Database equipped with the PGVector plugin. This database configuration allows for the storage of original text chunks alongside their respective embedding vectors in a distinct vector column \footnote{PGVector: \url{https://github.com/pgvector/pgvector} (Accessed: 28.02.2024)}. The embedding vectors are derived using an OpenAI embedding model \footnote{OpenAIEmbedding: \url{https://platform.openai.com/docs/guides/embeddings} (Accessed: 28.02.2024)}. The vector storage can be used to retrieve useful information during the query stage. The process is illustrated above in Fig. \ref{fig_2_rag}, where a search for a user's query is started within the vector storage, retrieving the $k$ most similar chunks for the given query (we experimented with different numbers of chunks and found good results with three chunks). These chunks are integrated as context into the prompt given to the Large Language Model (LLM).

One significant drawback of OpenAI’s models is the fact, that they are close-sourced and operated by OpenAI which might pose privacy issues and, at least early-on in the project, excluded the possibility of further training on specific data as envisioned in our project. In our case, we therefore used as an additional model Llama-2-13B  from Meta as it is openly accessible and can be further trained locally \citep{touvron2023llama,touvron2023llama2}. Meta's LLaMA series of models was trained only on publicly available data (overall around 1.25 trillion tokens, the majority of data is gained from CommonCrawl with about 850 billion tokens). This model was used in different setups: a baseline approach in which the model was directly queried and, again, a setup in which the prompt was enriched using the system message as well as RAG (prompt engineering and RAG are identical to the approach used with GPT-3.5). 

\begin{table*}[tbh]
\centering
\ra{1.1}
\captionsetup{format=hang}

\caption{Hyperparameters selected for Fine-Tuning the LLaMA-2 model (13B parameters).}

\begin{tabular}{l|c}
\toprule
\parbox[l]{5cm}{\centering Parameter} & \parbox[l]{8cm}{\centering Value} \\ 
\midrule
Optimizer & paged\_adamw\_32bbit \\ 
Learning rate & 1e-4 \\ 
LR scheduler type & cosine \\ 
Warmup ratio & 0.1 \\ 
Max grad norm & 0.3 \\ 
Epochs & up to 128000 \\ 
\bottomrule
\end{tabular}

\label{tab:fine-tuning}
\end{table*}

As one further step, LLaMA-2 was used in a fine-tuned version. The model was hosted and trained locally (we used a workstation equipped with a NVIDIA RTX A6000 with 48 GB of RAM on the graphics card). Therefore, we decided to use the mid-size LLaMA for all our experiments (with 13 billion parameters) as this still allowed to train and fine-tune the model when using quantization and LoRA \citep{hu2021lora}. For further details on training, see table \ref{tab:fine-tuning}.

\subsection{Evaluation of LLMs in Question-Answering}
Measuring question-answering is difficult because for many questions there are multiple correct answers, e.g.,
two answers to a question might simply be formulated quite differently, using synonyms, or having a different sentence structure. None-the-less, both answers might be correct. This makes comparison to a ground truth difficult. There are different approaches for question-answering and quite a large number of metrics have been proposed. A couple of these focus on overlap between the model's output and the ground truth response provided by a human. BLEU-4 is widely used as such a metric \citep{papineni2002bleu} providing values in the range of $0$ to $1$. Furthermore, we also use ROUGE-1 \citep{lin2004rouge} and METEOR \citep{banerjee2005meteor} for evaluation which have been found to best correlate with human evaluation \citep{kim2020beyond}. As a further more modern metric, we in addition computed the BERTScore \citep{zhang2019bertscore} which again is calculating the similarity between token in the output by the model compared to the ground truth. Importantly, while other metrics count exact matches, the BERTscore computes semantic similarity for token in the contextual embedding space. 

It is currently assumed that such metrics by itself are not sufficient to analyze LLMs \citep{maynez-etal-2020-faithfulness}. Therefore, human evaluations are still used as a gold standard for assessing the capabilities. As a first step, we setup a web-based framework that present a question together with an answer to a user. The user is tasked with judging how helpful and how correct the given answer is. There is additional information provided for each question (the original slide for which we generated the question, plus a context of neighboring slides). This human evaluation has only been tested with two teaching assistants for the given robotics course which can be considered domain experts (detailed results are shown in the Appendix in table \ref{table_human_eval}). We will run this evaluation in the upcoming semester inside the robotics course.

Using another LLM---and often a more capable one with larger parameter counts---is currently an active area of research. Approaches as LLMEval \citep{lin2023llmeval} use an LLM for similarity rating itself, often using an elaborate prompt that guides the model to mimic a human-like evaluation.  In general, LLMEval showed some promising results, but appears difficult to directly apply in our approach. In our case, the integration of RAG lead to quite long context inputs which has shown problematic for LLMEval. Therefore, we adapted their general approach and prompted an LLM (GPT-3.5) as an evaluator which was provided with the original question, the generated answer, and the ground truth answer. In particular, we developed three different prompts that addressed different parts of our evaluation. First, `GPTSimilarity' is an evaluation of the similarity between the generated response and the ground-truth response in the test set. Secondly, for the metrics of the human evaluation, we introduced `GPTRater' which evaluated independently trustworthiness and helpfulness on the same scale as given to human evaluators (details see below).




\begin{figure}[htb]
    \centering
    \parbox{\columnwidth}{
        \centering
        \includegraphics[width=\columnwidth]{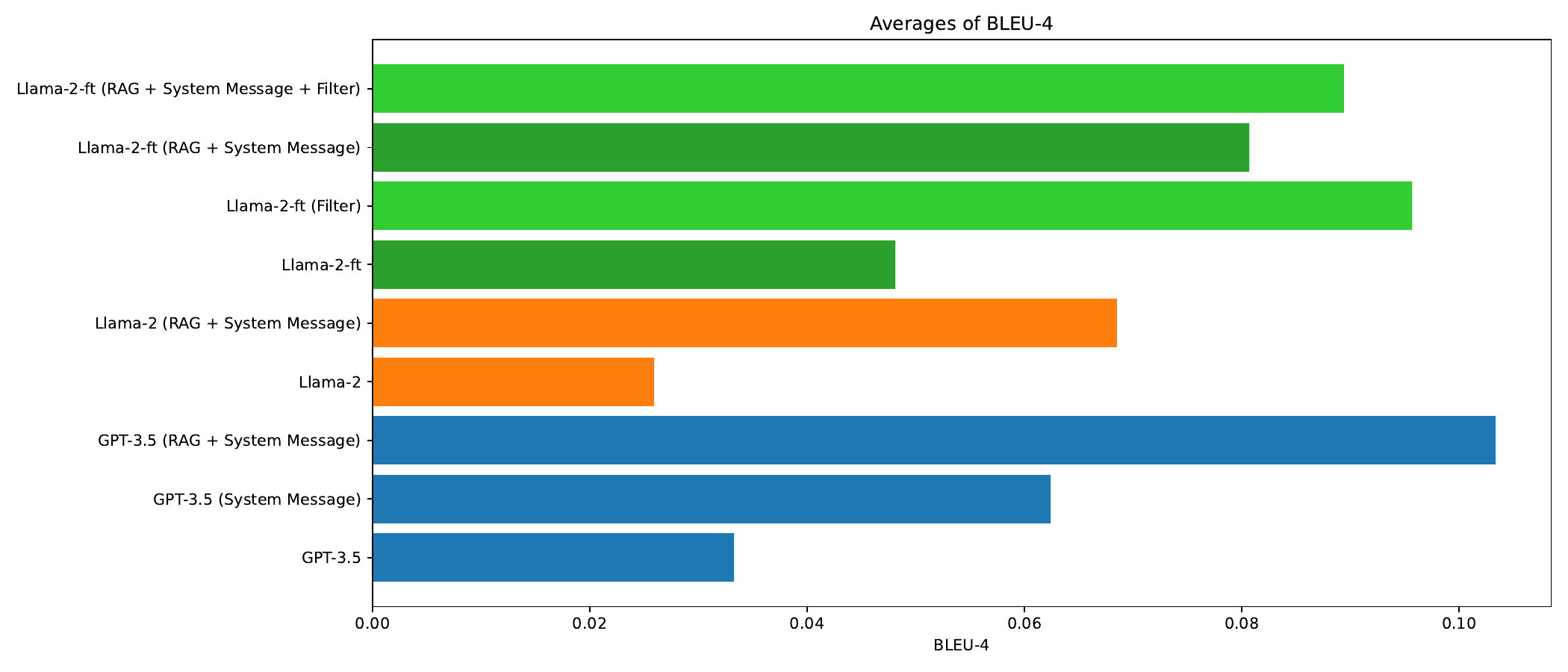}
        \centerline{(a) BLEU-4 Score}
    }
    \vspace{1em} 
    \parbox{\columnwidth}{
        \centering
        \includegraphics[width=\columnwidth]{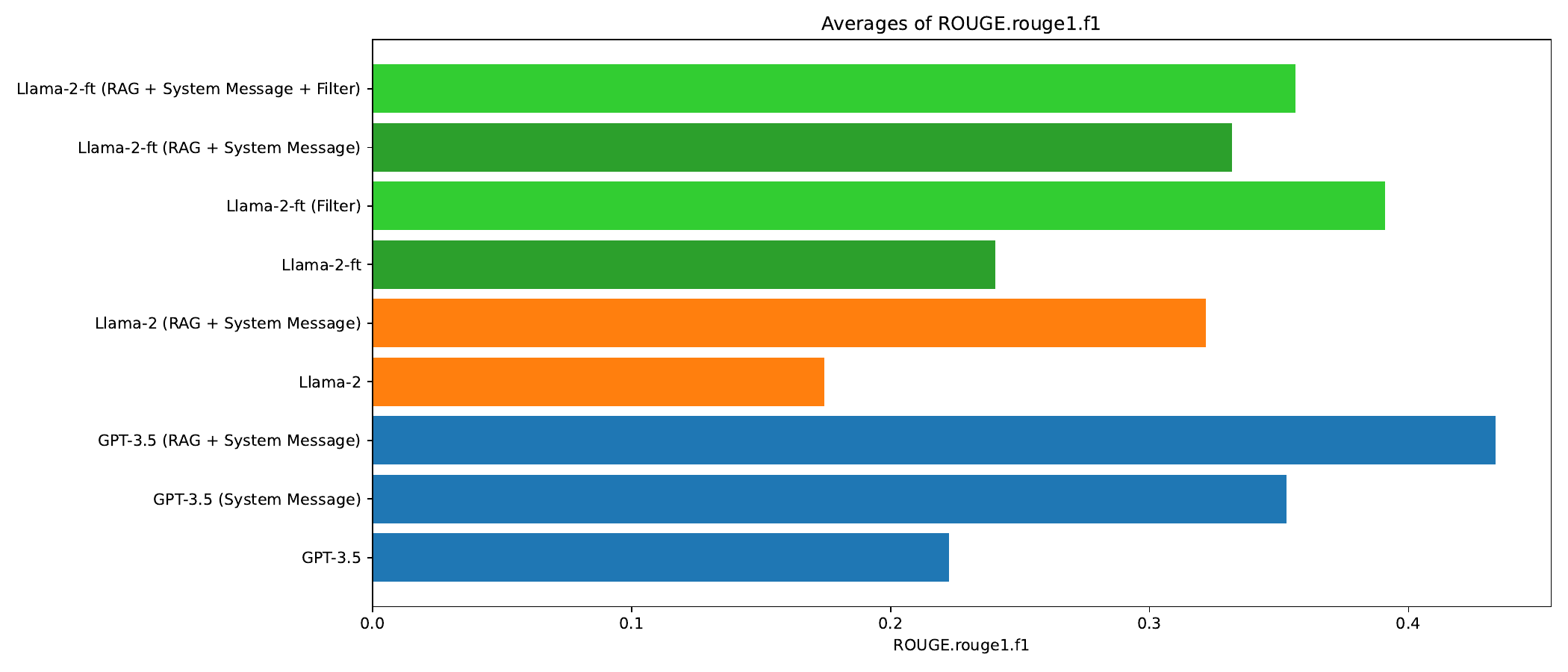}
        \centerline{(b) ROUGE Score}
    }
    \caption{Evaluation of Large Language Models: (a) BLEU-4 score which measures the precision of n-grams (here 4-grams) in the generated text compared to the ground truth text, while (b) ROUGE evaluates recall, measuring the overlap of n-grams between the generated and reference texts.}
    \label{fig_bleu}
\end{figure}

\section{Results} 

We tested the different LLM versions and different applied techniques in a question answering task with the questions tailored to the robotics course from the test set. As a first model, we used ChatGPT (GPT-3.5) as a standard high-performing and well-known general model. For GPT-3.5 we could apply prompt engineering and RAG as we accessed the model through OpenAI's API and simply enriched the prompt. As we considered fine-tuning as a third step, we also evaluated an open model in LLaMA-2. This lead to multiple different setups and different groupings: On the one hand, we used two different models of quite different size (GPT-3.5 has around 175 billion parameters, LLaMA-2 in the midsize version of around 13 billion parameters). On the other hand, we used different input versions for both models: Simply providing a question as an input, prompt engineering using an optimized prompt that provides context on the use case, and, in addition, Retrieval-Augmented-Generation in which the input was further enriched using background information from the lecture as additional input to the model. In the case of LLaMA-2 we further used fine-tuning and trained the model on a training dataset of example questions and answers for the robotics course.

\subsection{Comparison of Different LLM Extension}
As a first research question, we are interested in how well do different models perform and how do the different extension techniques affect the performance. Therefore, we evaluated these models on our test dataset. The different metrics compare the provided ground truth answer from the test set with the answer given by the model on the particular question (for detailed results see Appendix \ref{app_overview}, table \ref{tab_overview_gpt} and table \ref{tab_overview_llama}). Given are the BLEU-4 and ROUGE similarity scores in Fig. \ref{fig_bleu}. As a first observation, for both generic LLM models (indicated by colors: GPT-3.5 in blue, LLaMA-2 in orange) extending the prompt always improved the response considerably. 
Just adding a simple system message that instructs the model to act as a tutoring system showed a considerable improvement (Fig. \ref{fig_bleu}, lower part, comparison of GPT-3.5 with GPT-3.5 (System Message) in which a general message on the role and topic was provided to the model).
Retrieval-Augmented-Generation and prompt engineering 
show a further improvement and quite large positive effect. 


For the fine-tuned LLaMA-2 (shown in darker green and lighter green) these results differ slightly: Importantly, in training on our specific dataset the model was fine-tuned on a specific interaction structure that further included additional keywords, e.g., start and stop symbols. After learning these, the model tended to overproduce some of such tags which could be easily cleaned up as these were not fitting to the predefined structure. For example, when the model produced an end-tag but still continued to produce output, the text following that end-tag could be dropped. Therefore, we filtered the output of the fine-tuned model (the filtered and cleaned-up responses are shown in the figures in light green). This improved the similarity scores dramatically which might be partially explained as the texts were significantly shortened 
(for the fine-tuned LLaMA-2 model output length dropped from $413\pm153$ words to $109\pm81$ words after filtering; when integrating RAG output length dropped from $246\pm181$ words to $173\pm109$ words after filtering).
There is an assumed bias of such similarity metrics for shorter answers (see below our analysis and comparison on relation between the different metrics). One curious observation after filtering was that RAG and prompt engineering didn't show an advantage for the fine-tuned version. This might be due to the training dataset to which the model fitted its interactions which didn't include RAG.

When considering fine-tuning, we compare the different versions of the LLaMA-2 model (all with 13 billion parameters), i.e., shown in orange and the fine-tuned version in green. Fine-tuning improved in all cases the results considerably, in particular when filtering the results afterwards. Last, when comparing the results of the fine-tuned model with GPT-3.5 (which is much larger with 175 billion parameters), we observed that a fine-tuned LLaMA-2 (with filtering) without additional prompting techniques performed considerably better as GPT-3.5. Adding RAG and prompting techniques (see above) improved GPT-3.5 by a huge margin and this model setup was the best we found in evaluation. But the margin towards fine-tuned LLaMA was slim and our fine-tuning training data appeared to have a detrimental effect to RAG which should be further analyzed (see note above).

\begin{figure}[t!b]
    \centering
        \centering
        \includegraphics[width=\columnwidth]{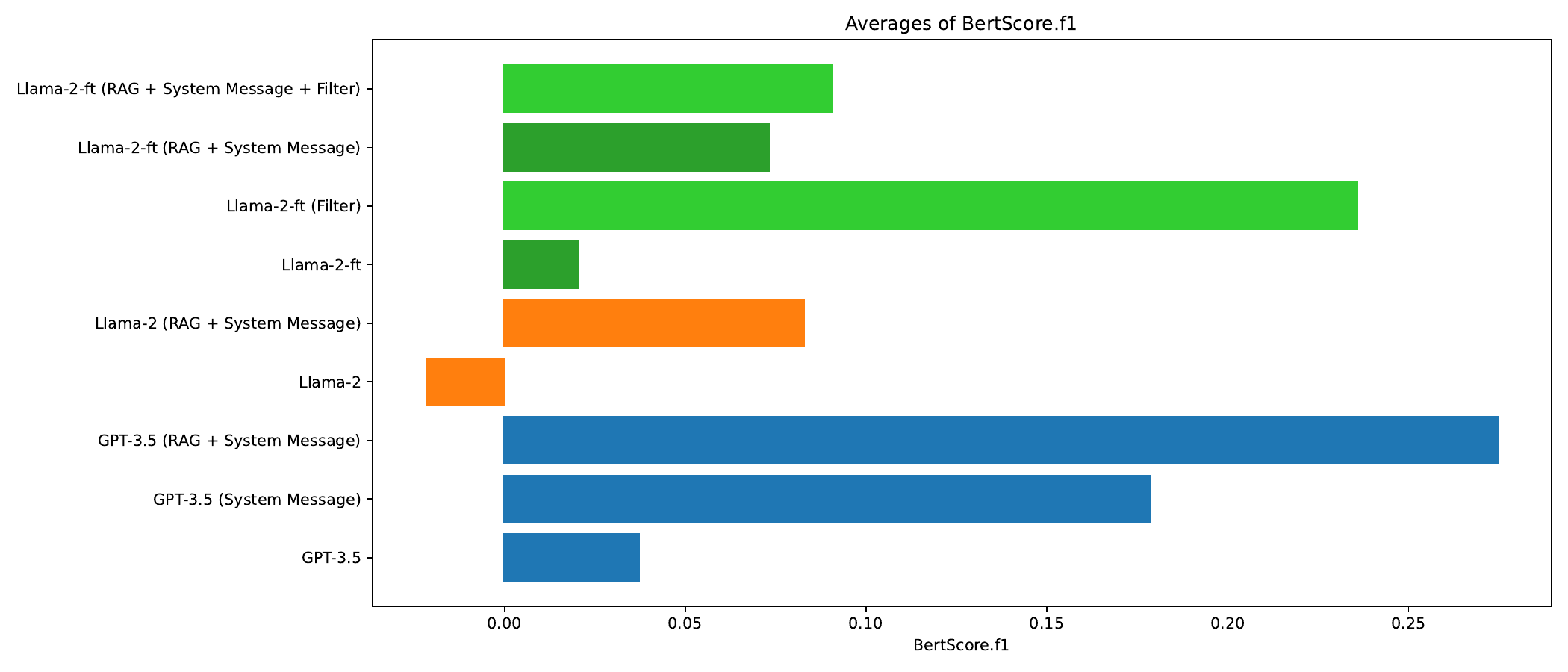}
	\caption{BERTScore as an Evaluation Metrics for Large Language Models which uses BERT embeddings for semantic similarity computation.}
    \label{fig_bert}
\end{figure}

For the METEOR score most models perform almost equally good (not shown in detail). Except for the base model with raw data and question only, which performs slightly worse. In BERTScore, we, first, observe as a similar pattern that applying RAG and using a specific prompt dramatically improves the measured similarity (Fig. \ref{fig_bert}, comparing inside the specific models---LLaMA-2 respectively GPT-3.5---the versions with RAG and without RAG). Secondly, again similar reflecting the results from above, fine-tuning (showing in green) introduces a strong advantage compared to the original model (shown in orange). Again, integrating RAG after fine-tuning appears problematic. What is noteworthy from the BERTScore evaluation is that numbers appear overall more pronounced. 


\begin{figure}[t!b]
\centering
\includegraphics[width=\columnwidth]{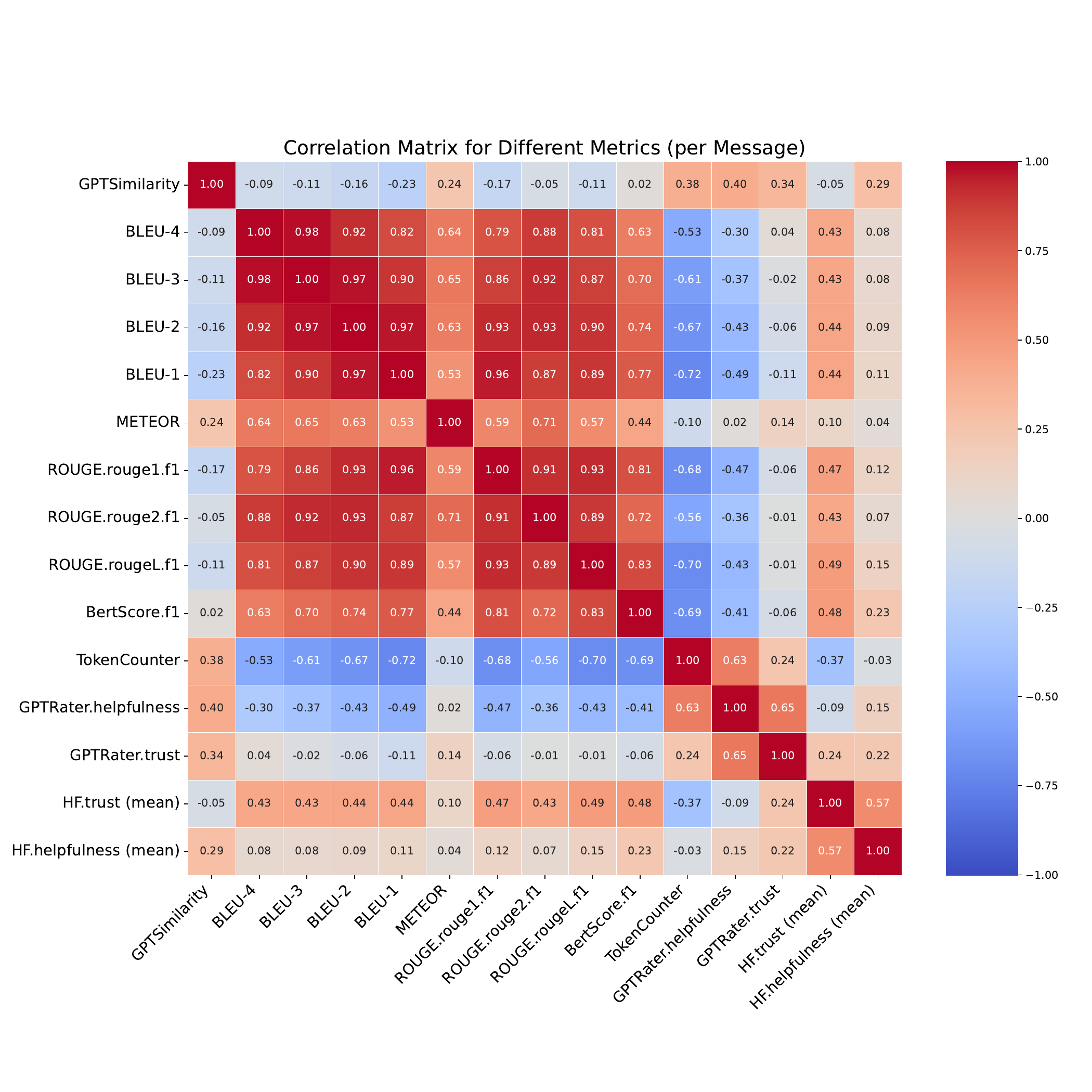}%
\caption{Correlation Matrix of Evaluation Metrics: The color-coded matrix represents correlations between evaluation metrics used to assess the performance of LLMs (measured on our test set). Each cell in the matrix indicates the correlation coefficient between two metrics, with color intensity (from blue to red) reflecting the strength of the correlation. Metrics include traditional similarity scores like BLEU and ROUGE, as well as newer ones like BERTScore, and their relationships with model output features, e.g., token count and human evaluation scores on trustworthiness or helpfulness.}
\label{fig_comp_correlation}
\end{figure}

\subsection{Analysis of Evaluation Metrics} 
Evaluation of LLMs and their responses is difficult and there are multiple metrics and different approaches have been proposed. We further analyzed some of these metrics: First, we considered how the different metrics correlate with each other in order to understand better what these represent and how they might capture similar characteristics. In addition, we computed correlations to other characteristic features of the given responses. As one example, we considered the length of the response as it has often be assumed that there is a bias for shorter answers in some of the more traditional metrics. 

We observed (Fig. \ref{fig_comp_correlation}) that the traditionally computed similarity metrics as BLEU, ROUGE, and BERTScore (to a slightly lesser extend) are highly correlated. Interestingly, all these metrics showed a negative correlation with the token count. In our case, we found the assumption confirmed that shorter sentences are preferred by these metrics.
METEOR also showed a positive---but smaller---correlation with the metrics above. It appeared nearly uncorrelated with the token counter (only a very small negative correlation). GPTSimilarity (results are shown in the appendix, Fig. \ref{app_gptsim}), as a measurement that prompts an LLM to evaluate similarity, did not show any large correlation with the more traditional similarity metrics, but surprisingly has a positive correlation with the token count, meaning in our evaluation longer answers were usually favored (this should be further analyzed).

Secondly, we want to understand the relation of the different metrics to a human evaluation (considered as the gold standard). The human evaluation should produce reproducible assessments based on recognizable criteria, ensuring that the evaluations are objective and can be consistently applied. 
Our approach for the human evaluation follows the method described in tolokan.ai \footnote{https://toloka.ai/blog/perplexity-case/}. Their evaluation process utilizes two main criteria: Helpfulness and Trustworthiness (which should reflect how well the answer is backed up by argument or references). Each measured on a scale of five answers (helpfulness: not helpful -- repetition -- unclear -- limited -- helpful; trustworthiness: nonsense -- false statement -- general knowledge -- partially proven --proven).
In the evaluation, each sentence of a model's response is evaluated individually, allowing for a granular analysis of the output. Importantly, in this evaluation method a weighting of each rating for a sentence is additionally computed considering the length. This weighting is crucial for determining the proportional impact of each rating level on the overall response. As a consequence, the length of the full answer is not a direct criteria for good or bad answers. For example, as long as the answer provides helpful information with respect to the request, a long answer can still be useful to answer the request in more detail. 
As a first step, the evaluation was conducted by only two teaching assistants who assessed a randomly selected subset of the test set ($130$ question-answer pairs). 
This of course limits the reliability of the human feedback scores as the number of participants is too small. The study shall be repeated in the future, as we want to use it in an upcoming semester in the robotics course. Still, we hope that we can gain first insights from the human feedback. It is of further note that the evaluation was not done by naive subjects, but by experts on the topics of the course. This should have a positive impact on their knowledge of the topics of the course. 
The human labelers were blindly presented responses from the different models and extensions and asked to label these for helpfulness and trustworthiness. Afterwards, this feedback was pooled together for each specific LLM configuration individually. 

As a first observation, when adding RAG into any model (GPT-3.5 or LLaMA-2) this increased trustworthiness in the model (results are not shown in detail). This is to be expected as base models---without any extension---are not capable of providing references to back up their answers. Adding references appears to have a direct positive impact on trustworthiness. 
For helpfulness we couldn't find a clear trend (e.g., GPT-3.5 already proves as a strong baseline in this respect which gives always answers that relate to the question). Overall, we are interested in how the human evaluation correlates with the computed metrics (Fig. \ref{fig_comp_correlation}). There is a good correlation of trustworthiness with BLEU, Rouge, and BERTScore (as a cautious reminder: the human evaluation is currently only restricted to feedback by two teaching assistants). Trustworthiness is also negatively correlated with the token count. This appears reasonable as the scale of trustworthiness rewards answers that are backed up by argument. In contrast, well known facts that might be known from training data are evaluated as neutral. There is no strong correlation for helpfulness with respect to the other metrics, only a small correlation with BERTScore. Trustworthiness and Helpfulness appear correlated. 

One interesting comparison is between the human feedback on trustworthiness and helpfulness compared to similar ratings obtained from an LLM---following the GPTRater approach which estimates trust and helpfulness on a given answer with respect to the ground truth answer using a LLM (scores for different LLM variations using the GPTRater approach, that evaluates trust and helpfulness using an LLM, are given in the appendix, Fig. \ref{app_gptrater}). For these we find small correlations, but again these should be handled carefully as long as there is only human feedback given by two participants.


\begin{figure}[t!bh]
    \centering
    \begin{subfigure}{\textwidth}
        \centering
        \includegraphics[width=0.65\columnwidth]{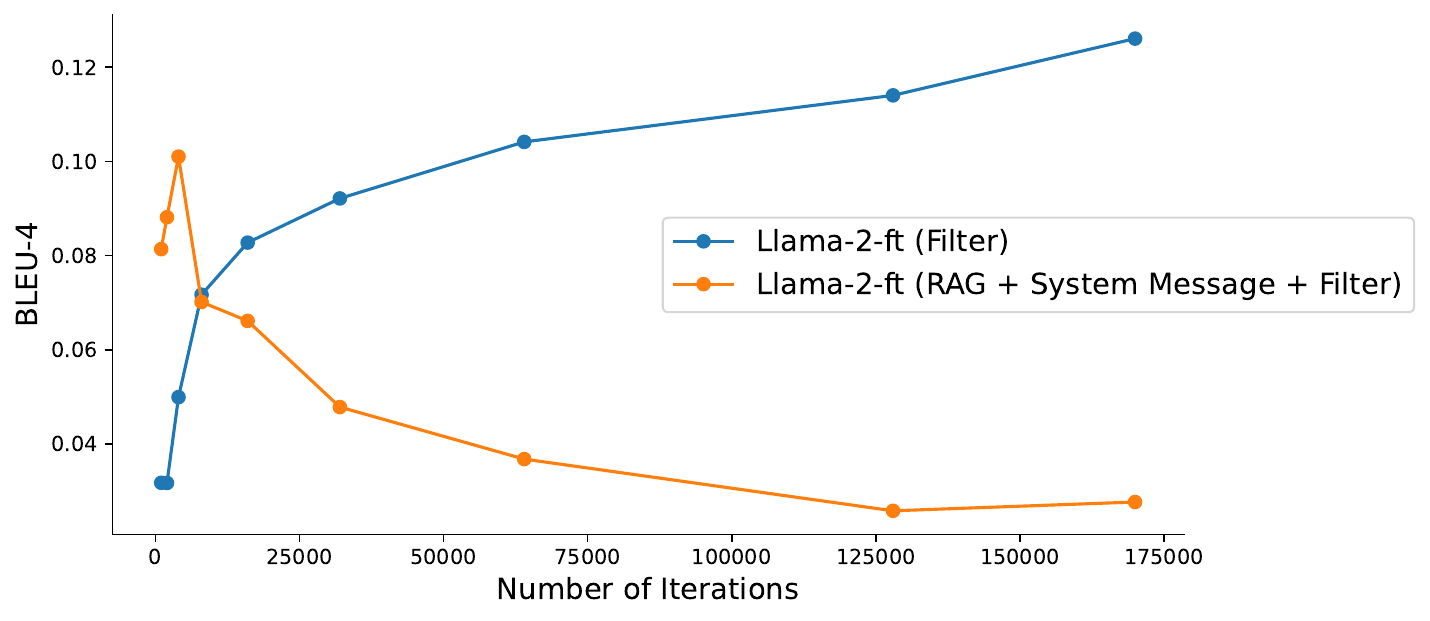}
        \caption{Development of BLEU-4 Score during training.}
        \label{fig:bleu4_train}
    \end{subfigure}
    \vfill
    \begin{subfigure}{\textwidth}
        \centering
        \includegraphics[width=0.65\columnwidth]{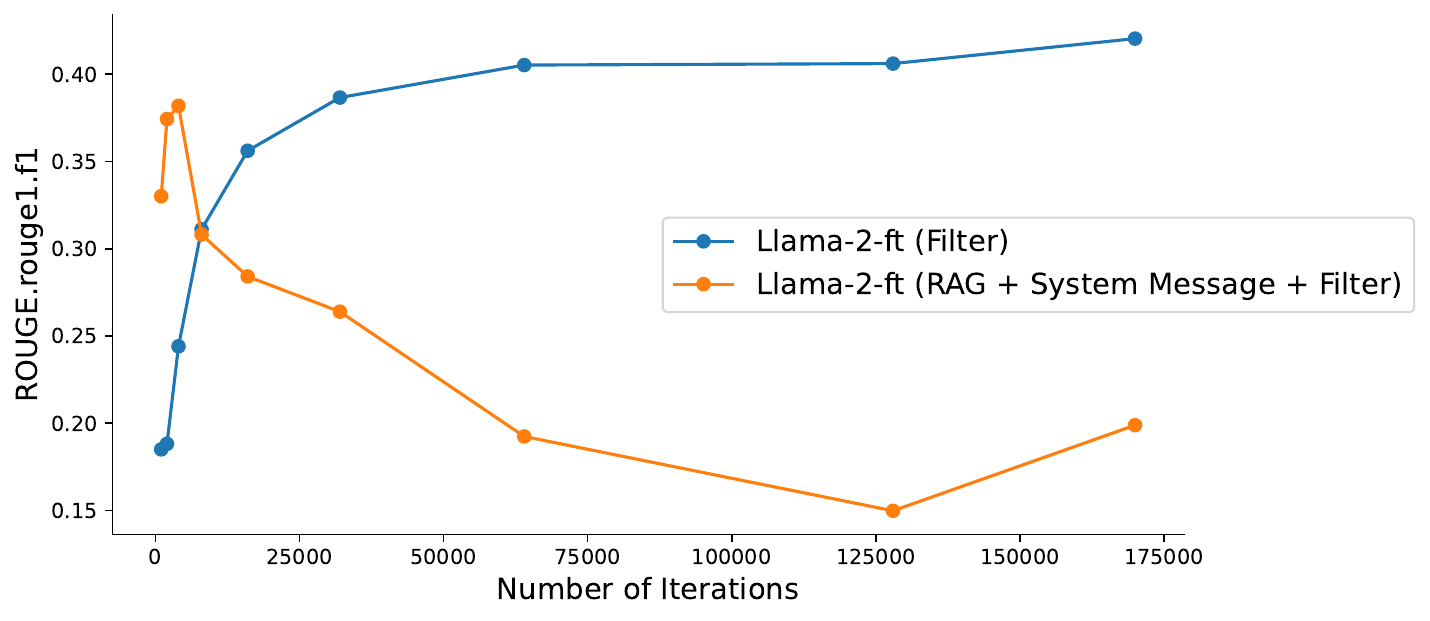}
        \caption{Development of ROUGE Score during training.}
        \label{fig:rouge_train}
    \end{subfigure}
    \caption{Evaluation of LLM during fine-tuning: Development of metrics measured for different checkpoints during the training process. (a) BLEU-4 score which measures the precision of n-grams (up to 4-grams) in the generated text compared to the ground truth text, while (b) ROUGE evaluates recall, measuring the overlap of n-grams between the generated and reference texts. Compared are two models, first, a simple LLaMA model and, secondly, a configuration that included RAG.}
    \label{fig_eval_training}
\end{figure}

\subsection{Adaptation of LLM during Fine-Tuning} 
During the fine-tuning of the model on the training dataset, we aimed to observe how the model's responses evolved over the course of training. Specifically, we investigated how long it took for the model to adapt to the specific task structure and at what point the model began to overfit to the training data. Therefore, we recorded model weights  
throughout the training process. A checkpoint was generated for every 250 training steps. 

For an analysis, we selected checkpoints exponentially spaced by a factor of $2$, examining model responses after $1000$, $2000$, $4000$, $8000$, $16000$, $32000$, $64000$, $128000$ training steps, and at the end of training ($178000$ steps). Multiple sample answers generated by the fine-tuned models were evaluated in response to questions from the test set. Given the issues encountered when applying Retrieval-Augmented Generation (RAG) to a fine-tuned model, we run this evaluation in both kind of settings: Using the fine-tuned model or in addition applying RAG on the fine-tuned model.

The selection of models was based, first, on human evaluations according to specifically defined criteria, including correctness of the answer, reference to the lecture, correct use of the specified formatting, the overall amount of formatting errors, and grammar.  Ultimately, checkpoints $2000$ and $128000$ were selected for a broader comparison in human evaluation. 

Despite the limited scope of the human evaluation, some general trends were observed. Checkpoint $128000$ yielded the best results for the `Fine-tuned LLaMA' model configuration when used without applying RAG. In contrast, already checkpoint $2000$ performed best for the model when used with RAG (`Fine-tuned LLaMA + RAG'). 
Example answers to a sample question are presented in the appendix (table \ref{tab_sample-question} shows the ground truth and the input question; Table \ref{tab_ft_comp} compares the fine-tuned models response without using RAG for the two different checkpoints, i.e., early and late in training; table \ref{tab_ft_comp_rag} compares the fine-tuned models response when additionally integrating RAG for the two different checkpoints). The performance of the checkpoints deteriorates when the configurations are switched. 

Secondly, we computed the different metrics for these two type of settings throughout the training process. 
Interestingly, there was a strong correspondence between the human assessments and the performance metrics at specific training points (see Fig. \ref{fig_eval_training}). For the plain LLaMA-2 model without RAG, the scores (e.g., BLEU-4 and ROUGE, shown in blue in Fig. \ref{fig_eval_training}) improved consistently over time and continued to improve late in training. In contrast, with RAG integration (shown in orange), fine-tuning peaked early and then dropped considerably. The metrics indicate an optimal performance around $4000$ training steps, after which the system's performance deteriorated drastically. This aligns well with the human assessment, which chose the model's responses after $2000$ training steps as optimal which is quite close and for which the model performed on a similar level when considering the metrics.

We assume that extended training in the RAG configuration leads to overfitting towards the expected target structure. 
It appears as an excessive over-adaptation to the input format which leads to overfitting and partial output of the system prompt. The model attempts to adhere strictly to this pattern, which might further cause hallucinations, although this requires further analysis and a more in-depth evaluation. Conversely, when considering the early stage of training---in our case checkpoint after $2000$ steps---the fine-tuned configuration without RAG did not produce satisfactory results, likely because it had not undergone sufficient fine-tuning to accurately replicate the desired formatting for references and lecture content.

\section{Discussion and Conclusion} 
The evaluation of a Large Language Model-based tutoring system for a University robotics course highlighted several insights into the application of advanced LLM techniques and the resulting performance in an educational setting. First, our findings underscored the positive impact of Retrieval-Augmented-Generation (RAG) and prompt engineering, which consistently improved model performance across similarity metrics. Particularly, the use of RAG demonstrated a considerable enhancement in providing factual answers and is consistent with the general belief that RAG is reducing hallucinations \citep{shuster-etal-2021-retrieval-augmentation}. Furthermore, even though our human evaluation is currently restricted to two test subjects, their answers already point out that added references increases trustworthiness. Therefore, RAG appears as a very valuable technique that should be---together with some form of prompt engineering---considered first. As further advantages, in our experience RAG is quite straight forward to realize, in particular in a course setting in which well-curated background material is readily available. Furthermore, from a teaching point of view a tutor should stick to the lecture material, e.g., when going over a concept the tutoring system should carefully choose examples and ideally stick---or at least start---with the ones provided in the lecture. This should positively affect the learning of students.

Fine-tuning has to be considered as a more involved technique. It requires additional effort in setting up a data set for training. As an advantage, in our case we saw that a quite small fine-tuned model (13 billion parameters) consistently performed on the same level---or better---as GPT-3.5 (175 billion parameters) when used without RAG. Fine-tuning produced a much more efficient expert which showed as quite capable. But, on the downside, the process of fine-tuning appeared as more delicate. In our data, we observed a curious drop-off when adding RAG to the fine-tuned model which was unexpected and would contradict our and others' experience with RAG. As an explanation, fine-tuning aims to specialize a model to a specific task and a specific type of interaction. A fine-tuned model might loose some of its general flexibility. As a consequence, when interacting very differently with the model, the model might produce worse results or even behave erratically. The introduction of RAG drastically changes the input to a model and this change appears as problematic when RAG is introduced after fine-tuning. Therefore, the combination of techniques should be considered carefully beforehand. In the future, we want to further test how fine-tuning on interactions that include RAG affect the result which, unfortunately, requires a different data set.

We further compared different metrics to evaluate question answering in LLMs and found a high correlation between BLEU-4, ROUGE, and BERTScore. The analysis of these metrics revealed an inherent preference for shorter responses, suggesting that these metrics might be biased against lengthier, albeit potentially more informative, answers. This observation is crucial for designing future LLM evaluation frameworks, as it challenges the effectiveness of current automated metrics in truly capturing response quality in educational settings.
LLM-based evaluations still appear difficult as we could not demonstrate a large correlation of a similarity evaluation with the more conventional metrics. With respect to trustworthiness and helpfulness, our human evaluation has to be based on a larger test group and we want to extend this in the future. 


\section*{Acknowledgements}
This research was supported by the `Institut für Angewandte Informatik' at the University of Münster through an individual grant. The article is an extended version of a paper presented at the International Workshop on AI in Education and Educational Research (AIEER) at ECAI-2024 (27th European Conference on Artificial Intelligence).



\bibliographystyle{unsrtnat}
\bibliography{literatur}

\newpage

\appendix
\section*{Appendix}

\section{Overview Detailed Results for Evaluation of Different LLMs}
\label{app_overview}

\begin{table*}[tbh]
\centering
\ra{1.1}
\captionsetup{format=hang}

\caption{Detailed overview of metrics for the different GPT-3.5 configurations, shown as a comparison between a full RAG including a system message as a prompt and simply using the question as an input. Best value is highlighted by green shading.}

\resizebox{\columnwidth}{!}{
\begin{tabular}{l|c|c}
\toprule
\textbf{Configuration} & \parbox[c]{5.5cm}{\centering \textbf{GPT-3.5 \\(RAG + System Message)}} & \parbox[c]{5.5cm}{\centering \textbf{GPT-3.5 \\ (Question Only)}} \\
\midrule
GPTSimilarity & \cellcolor{graygreen} 89.95 $\pm$ 16.57 & 88.81 $\pm$ 16.83 \\
\midrule
BLEU-4 & \cellcolor{graygreen} 0.10 $\pm$ 0.07 & 0.03 $\pm$ 0.02 \\
BLEU-3 & \cellcolor{graygreen} 0.13 $\pm$ 0.07 & 0.05 $\pm$ 0.02 \\
BLEU-2 & \cellcolor{graygreen} 0.18 $\pm$ 0.08 & 0.07 $\pm$ 0.03 \\
BLEU-1 & \cellcolor{graygreen} 0.26 $\pm$ 0.09 & 0.11 $\pm$ 0.04 \\
\midrule
METEOR & \cellcolor{graygreen} 0.35 $\pm$ 0.10 & 0.28 $\pm$ 0.06 \\
\midrule
Rouge1.f1 & \cellcolor{graygreen} 0.43 $\pm$ 0.10 & 0.22 $\pm$ 0.06 \\
Rouge2.f1 & \cellcolor{graygreen} 0.21 $\pm$ 0.09 & 0.09 $\pm$ 0.04 \\
RougeL.f1 & \cellcolor{graygreen} 0.29 $\pm$ 0.09 & 0.15 $\pm$ 0.04 \\
\midrule
BertScore.f1 & \cellcolor{graygreen} 0.27 $\pm$ 0.12 & 0.04 $\pm$ 0.10 \\
\midrule
TokenCounter & 197.89 $\pm$ 88.52 & 420.01 $\pm$ 99.85 \\
\midrule
GPTRater.helpfulness & 84.98 $\pm$ 14.72 & \cellcolor{graygreen} 94.21 $\pm$ 4.58 \\
GPTRater.trust & \cellcolor{graygreen} 85.54 $\pm$ 11.39 & 83.88 $\pm$ 8.10 \\
\bottomrule
\end{tabular}
}

\label{tab_overview_gpt}
\end{table*}

\begin{table*}[tbh]
\centering
\ra{1.1}
\captionsetup{format=hang}

\caption{Detailed overview of metrics for the different LLaMA-2 (13B parameters) configurations. Shown is the comparison: First, between the base model and fine-tuned versions. Further-more, for all models we compare directly asking the model the question or enriching the input using RAG and a system message. In case of the fine-tuned versions, we further filtered the output.
}

\hspace*{-1cm}
\resizebox{18cm}{!}{
\begin{tabular}{l|c|c|c|c|c|c}
\toprule
\textbf{Model Type} & \multicolumn{4}{c|}{\textbf{Fine-Tuned}} & \multicolumn{2}{c}{\textbf{Base Model}} \\
\cmidrule{2-7}
\textbf{LLM Output}& \multicolumn{2}{c|}{\textbf{Filtered*}} & \multicolumn{2}{c|}{\textbf{Raw}} & \multicolumn{2}{c}{\textbf{Raw}} \\
\cmidrule{2-7}
\textbf{Prompting Type} & \parbox[c]{2.5cm}{\centering RAG + \\ System Message} & \parbox[c]{2.5cm}{\centering Question \\ Only} & \parbox[c]{2.5cm}{\centering RAG + \\ System Message} & \parbox[c]{2.5cm}{\centering Question \\ Only} & \parbox[c]{2.5cm}{\centering RAG + \\ System Message} & \parbox[c]{2.5cm}{\centering Question \\ Only} \\
\midrule
GPTSimilarity & 72.79 $\pm$ 27.96 & 59.10 $\pm$ 29.56 & 72.85 $\pm$ 28.12 & 55.00 $\pm$ 28.42 & 76.07 $\pm$ 24.03 & \cellcolor{graygreen} 78.02 $\pm$ 22.00 \\
\midrule
BLEU-4 & 0.09 $\pm$ 0.07 & \cellcolor{graygreen}  0.10 $\pm$ 0.08 & 0.08 $\pm$ 0.07 & 0.05 $\pm$ 0.04 & 0.07 $\pm$ 0.06 & 0.03 $\pm$ 0.02 \\
BLEU-3 & 0.11 $\pm$ 0.07 & \cellcolor{graygreen}  0.12 $\pm$ 0.08 & 0.10 $\pm$ 0.07 & 0.06 $\pm$ 0.05 & 0.09 $\pm$ 0.06 & 0.04 $\pm$ 0.02 \\
BLEU-2 & 0.15 $\pm$ 0.08 & \cellcolor{graygreen}  0.17 $\pm$ 0.09 & 0.13 $\pm$ 0.08 & 0.09 $\pm$ 0.06 & 0.12 $\pm$ 0.07 & 0.05 $\pm$ 0.02 \\
BLEU-1 & 0.22 $\pm$ 0.10 & \cellcolor{graygreen} 0.27 $\pm$ 0.10 & 0.20 $\pm$ 0.11 & 0.14 $\pm$ 0.08 & 0.18 $\pm$ 0.09 & 0.09 $\pm$ 0.03 \\
\midrule
METEOR & \cellcolor{graygreen} 0.30 $\pm$ 0.11 & 0.29 $\pm$ 0.10 & 0.30 $\pm$ 0.10 & 0.29 $\pm$ 0.07 & 0.29 $\pm$ 0.09 & 0.24 $\pm$ 0.05 \\
\midrule
Rouge1.f1 & 0.36 $\pm$ 0.12 & \cellcolor{graygreen} 0.39 $\pm$ 0.10 & 0.33 $\pm$ 0.12 & 0.24 $\pm$ 0.10 & 0.32 $\pm$ 0.11 & 0.17 $\pm$ 0.05 \\
Rouge2.f1 & \cellcolor{graygreen} 0.17 $\pm$ 0.09 & \cellcolor{graygreen} 0.17 $\pm$ 0.08 & 0.15 $\pm$ 0.09 & 0.10 $\pm$ 0.05 & 0.14 $\pm$ 0.08 & 0.07 $\pm$ 0.03 \\
RougeL.f1 & 0.24 $\pm$ 0.09 & \cellcolor{graygreen} 0.26 $\pm$ 0.09 & 0.23 $\pm$ 0.09 & 0.17 $\pm$ 0.07 & 0.22 $\pm$ 0.08 & 0.12 $\pm$ 0.03 \\
\midrule
BertScore.f1 & 0.09 $\pm$ 0.49 & \cellcolor{graygreen} 0.24 $\pm$ 0.27 & 0.07 $\pm$ 0.18 & 0.02 $\pm$ 0.14 & 0.08 $\pm$ 0.14 & -0.02 $\pm$ 0.08 \\
\midrule
TokenCounter & 173.16 $\pm$ 109.48 & 108.78 $\pm$ 81.13 & 245.83 $\pm$ 180.82 & 412.92 $\pm$ 153.10 & 252.67 $\pm$ 144.88 & 503.85 $\pm$ 49.69 \\
\midrule
GPTRater.helpfulness & 64.96 $\pm$ 27.25 & 63.11 $\pm$ 24.70 & 66.95 $\pm$ 26.95 & 62.08 $\pm$ 25.43 & 77.90 $\pm$ 22.10 & \cellcolor{graygreen} 92.83 $\pm$ 4.99 \\
GPTRater.trust & 64.47 $\pm$ 29.30 & 58.81 $\pm$ 25.38 & 67.62 $\pm$ 29.47 & 55.19 $\pm$ 27.20 & \cellcolor{graygreen} 77.78 $\pm$ 22.41 & 76.79 $\pm$ 13.39 \\
\bottomrule
\end{tabular}
}
\label{tab_overview_llama}
\end{table*}

\clearpage

\section{GPTSimilarity}

\begin{figure}[!h]
\centering
\includegraphics[width=\textwidth]{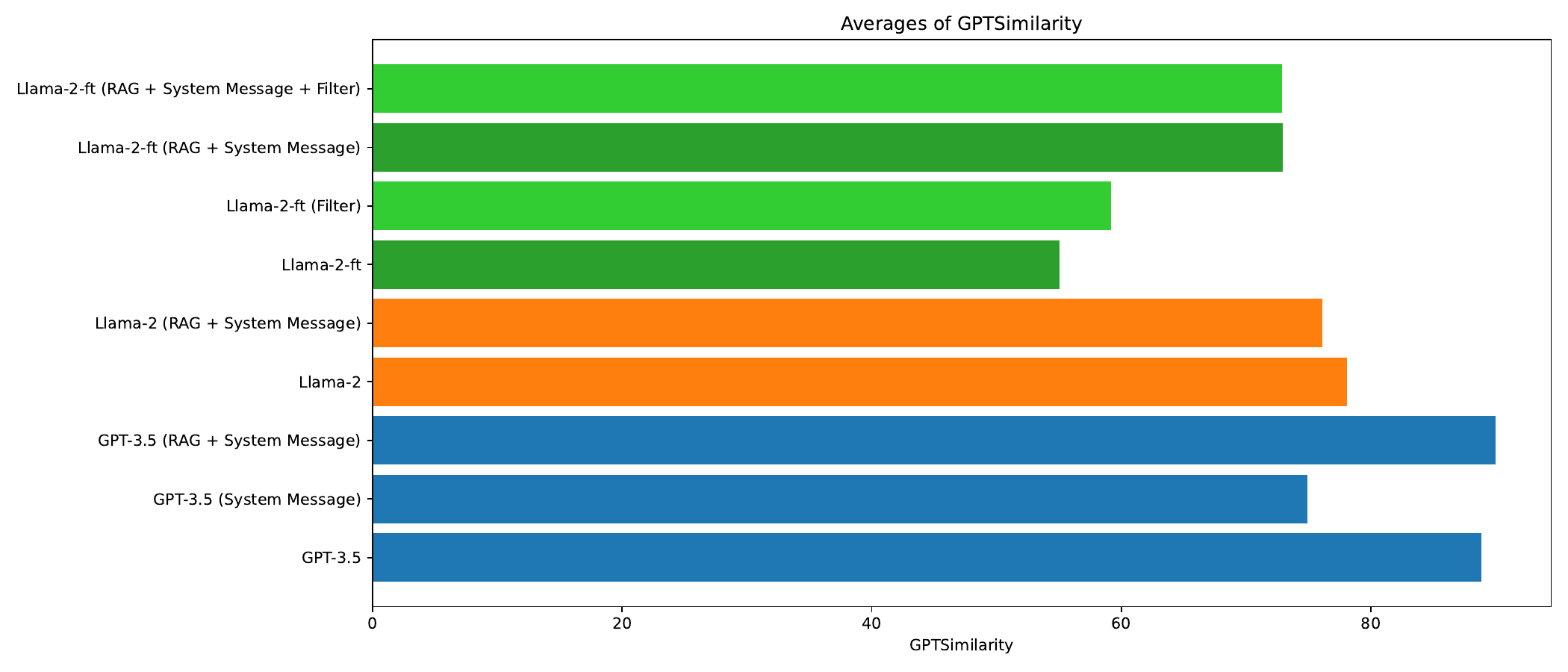}%
\caption{Evaluation of different LLM variations: GPTSimilarity Score on question answering is evaluated as a score by an LLM (GPT-3.5) that is specifically prompted for this task.
}
\label{app_gptsim}
\end{figure}

\clearpage

\section{GPTRater}

\begin{figure}[!h]
    \centering
    \begin{subfigure}{\columnwidth}
        \centering
        \includegraphics[width=\columnwidth]{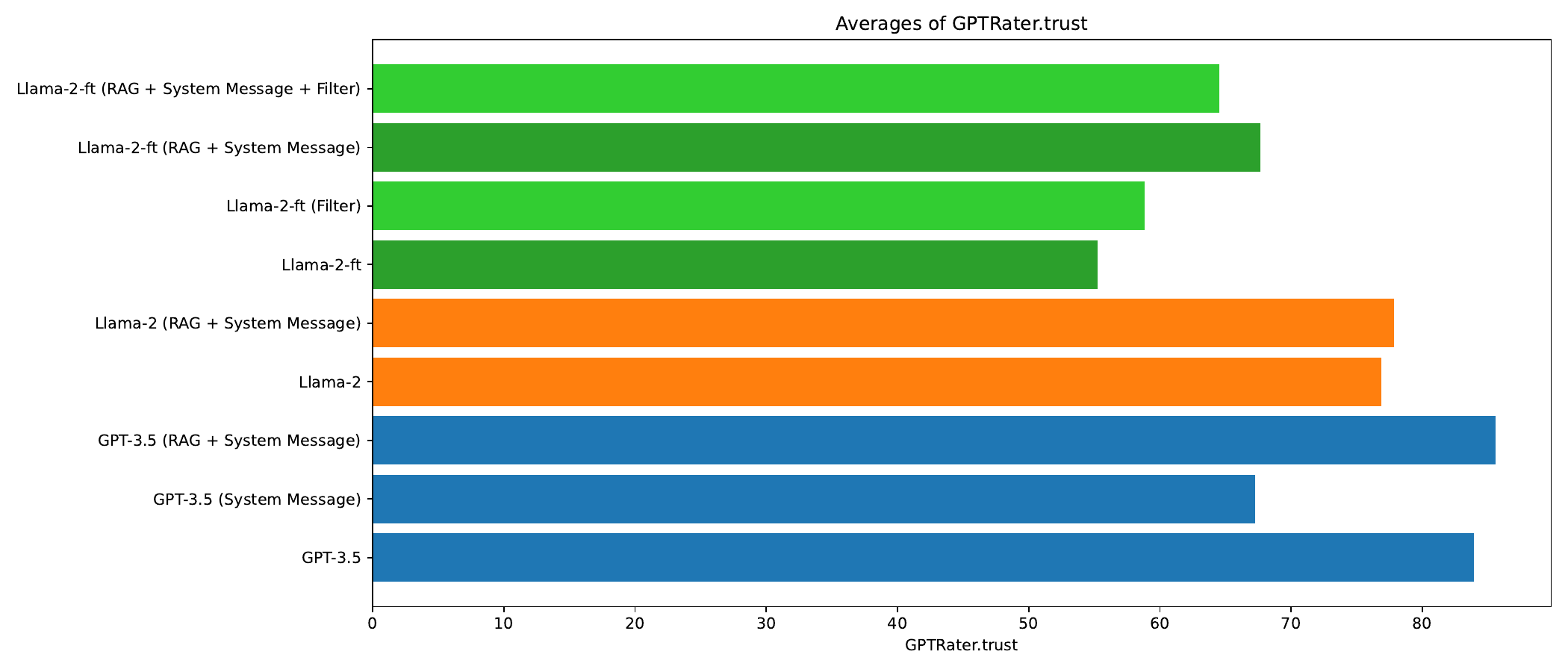}
        \caption{GPTRater.Trust Score}
    \end{subfigure}
    \vfill
    \begin{subfigure}{\columnwidth}
        \centering
        \includegraphics[width=\columnwidth]{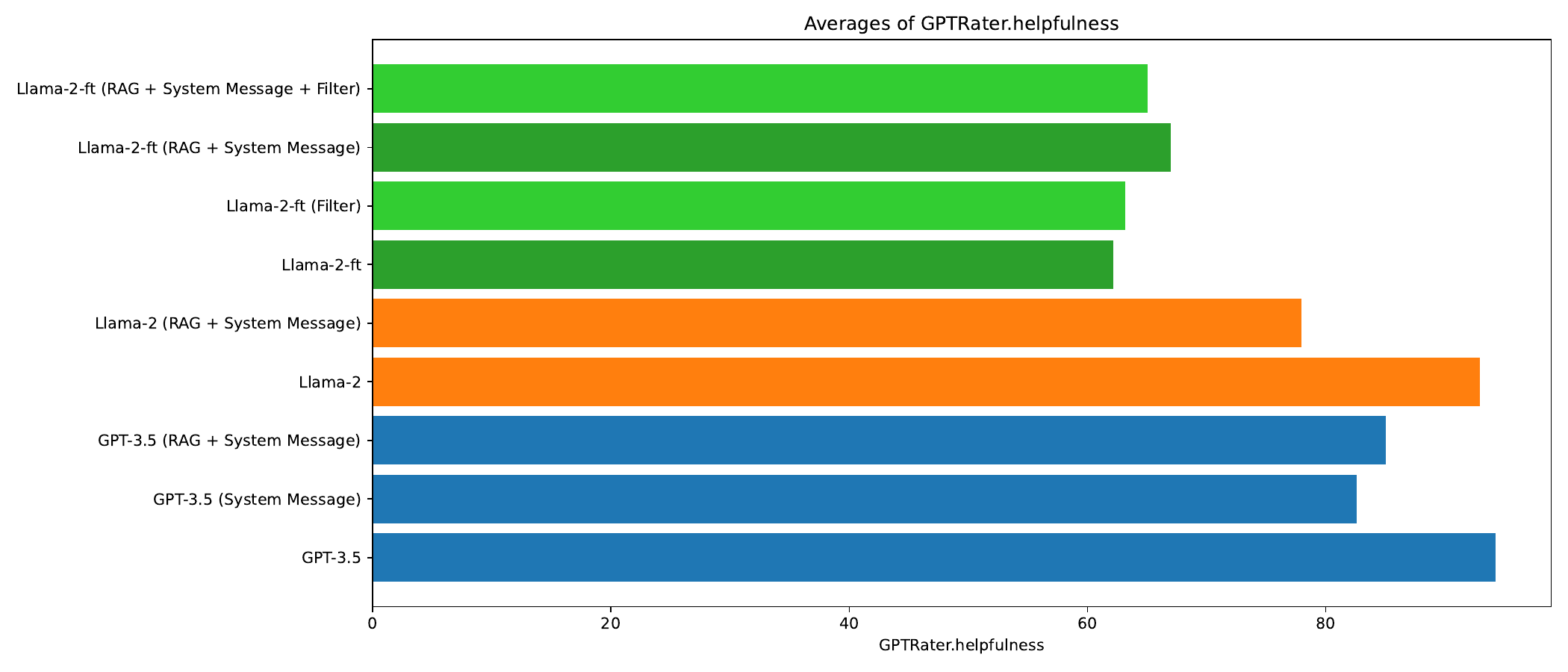}
        \caption{GPTRater.Helpfulness Score}
    \end{subfigure}
\caption{Evaluation of different LLM variations: GPTRater scores on question answering are evaluated as a score by an LLM (GPT-3.5) that is specifically prompted on an estimate for trust and helpfulness.
}
\label{app_gptrater}
\end{figure}

\clearpage

\section{Comparison of Responses during Fine-Tuning}

\begin{table*}[tbh]
\centering
\ra{1.1}
\captionsetup{format=hang}

\caption{Example pair of question together with the Ground Truth answer.}

\begin{tabular}{p{0.5\textwidth} | p{0.5\textwidth}}
\toprule
\textbf{Sample question} & \textbf{Ground Truth answer} \\ 
\midrule
\texttt{\textcolor{blue}{Provide an overview of different types of legal moves or operators that can be used in pathfinding algorithms for robot navigation.}} & \texttt{\footnotesize In pathfinding algorithms for robot navigation, legal moves or operators are typically described as possible atomic moves of the robot that are collision-free \textcolor{Green}{(@10-slam-deck Slide 11)} such as single steps between grid cells with four or eight connectivity \textcolor{Green}{(@10-slam-deck Slide 12)}. These legal moves must use an abstracted form of actions as primitives and might be associated with a cost \textcolor{Green}{(@10-slam-deck Slide 10)}. Different planning strategies may also involve increasing the size of obstacles by considering the size of the robot when determining legal moves \textcolor{Green}{(@10-slam-deck Slide 12)}.} \\ 
\bottomrule
\end{tabular}

\label{tab_sample-question}
\end{table*}


\begin{table*}[tbh]
\centering
\ra{1.1}
\captionsetup{format=hang}

\caption{Comparison of responses from different fine-tuned LLaMA-2 versions (after $2000$ and $128000$ steps), without using RAG.}

\begin{tabular}{p{0.5\textwidth} | p{0.5\textwidth}}
\toprule
\parbox[l]{7cm}{Fine-tuned LLaMA-2, \\ \textbf{2000 Training Steps}} & \parbox[l]{7cm}{Fine-tuned LLaMA-2, \\ \textbf{128000 Training Steps} } \\ 
\midrule
\texttt{\footnotesize Pathfinding algorithms for robot navigation often rely on various legal moves or operators to determine the sequence of actions that a robot can take to reach a goal. These legal moves  that the robot does not violate any constraints or obstacles in its environment. Here's an overview of different types of legal moves or operators commonly used in pathfinding algorithms for robot navigation: \textbackslash n\textbackslash n 1. Forward Movement (F): This operator allows the robot to move forward in a straight line, without any turns or [...]} & \texttt{\footnotesize \textcolor{Green}{[RESP]} In the context of robot navigation and pathfinding, different types of legal moves or operators can refer to various options for allowing a navigation algorithm to consider different types of actions while planning a path for a robot. These may include options such as forward, backward, turn left or right, combine movements, etc. The specific types of legal moves available will depend on the particular algorithm being used and the environment being navigated \textcolor{Green}{(@09-Localization-deck Slide 52)}. \textcolor{Green}{[/RESP]}} \\ 
\bottomrule
\end{tabular}

\label{tab_ft_comp}
\end{table*}

\begin{table*}[tbh]
\centering
\ra{1.1}
\captionsetup{format=hang}

\caption{Comparison of responses from different fine-tuned LLaMA-2 versions (after $2000$ and $128000$ steps) when enriching the input using RAG and prompt engineering.}

\begin{tabular}{p{0.5\textwidth} | p{0.5\textwidth}}
\toprule
\parbox[l]{7cm}{Fine-Tuned LLaMA-2 + RAG and \\ System Message, \textbf{2000 Steps}} & \parbox[l]{7cm}{Fine-Tuned LLaMA-2 + RAG and \\ System Message, \textbf{128000 Steps}} \\ 
\midrule
\texttt{\footnotesize Based on the lecture material, different types of legal moves or operators that can be used in pathfinding algorithms for robot navigation include:\textbackslash n\textbackslash n * Operators: Describe legal moves from state to state using an abstracted form of actions as primitives. \textcolor{Green}{(@Slide 10-slam-deck 10)}\textbackslash n * Operators: Explain how to control a mobile robot in order to move the robot towards some well-specified target using locomotion. \textcolor{Green}{(@Slide 09-Localization-deck 47)}\textbackslash n * Operators: Guide forms of [...] } & \texttt{\footnotesize \textbackslash n \textcolor{Red}{\#\#\# The lecture material and your chat with the student. Now, do the following steps:\textbackslash n\textbackslash n Requests: Based on your latest message, understand what the student wants to know by listing the request/requests based on your latest chat message.\textbackslash n Evaluate whether the request is understandable and can be answered with the lecture material above or not.\textbackslash n Use this format for referencing slides or Images by always incorporating the slide number and the lecture Title: (@Slide <name of} [...]} \\ 
\bottomrule
\end{tabular}

\label{tab_ft_comp_rag}
\end{table*}

\clearpage

\section{Overview human evaluation}

\begin{table*}[tbh]
\centering
\ra{1.1}
\captionsetup{format=hang}

\caption{Human evaluation of the different LLM versions (evaluation was conducted by only two teaching assistants who assessed a randomly selected subset of the evaluation set). Each sentence of an answer was presented to human labelers together with context. The labelers were tasked to rate this along two dimensions: Trust and Helpfulness. Each dimension was given on a five step scale. Trust: Nonsense -- False statement -- General knowledge -- Partially proven and shown -- Proven (with proof / reference in answer); Helpfulness: Not helpful -- Repetition -- Unclear -- Limited extent helpful -- Helpful.}

\resizebox{\textwidth}{!}{
\begin{tabular}{l|cc|cc|ccc}
\toprule
\parbox[l]{3cm}{\centering Class} & \parbox[c]{1.5cm}{\centering GPT-3.5} & \parbox[c]{1.5cm}{\centering LLaMA-2} & \parbox[c]{1.5cm}{\centering GPT +RAG} & \parbox[c]{1.5cm}{\centering LLaMA +RAG} & \parbox[c]{1.8cm}{\centering fine-tuned LLaMA} & \parbox[c]{1.8cm}{\centering LLaMA-ft +RAG} \\ 
\midrule
\multicolumn{7}{l}{\textbf{Trust}} \\
\midrule
\ \ Nonsense           & 0           & 0.005       & 0              & \textbf{0.165} & 0              & 0.095 \\ 
\ \ False Statement     & 0           & 0.015       & \textbf{0.035} & 0.03           & 0.015          & 0.03  \\ 
\ \ General Knowledge & 0.04        & \textbf{0.175} & 0.005        & 0.135          & 0.055         & 0.125 \\ 
\ \ Partial             & \textbf{0.95} & 0.795      & 0.05           & 0.12           & 0.275          & 0.22  \\ 
\ \ Proven              & 0           & 0           & \textbf{0.905} & 0.53           & 0.655          & 0.515 \\ 
\midrule
\multicolumn{7}{l}{\textbf{Helpfulness}} \\
\midrule
\ \ Not Helpful         & 0.025       & 0.02        & 0.035          & \textbf{0.25}  & 0.105          & 0.175 \\ 
\ \ Repetition          & 0.015       & 0.055       & 0.07           & \textbf{0.12}  & 0              & 0.07  \\ 
\ \ Unclear             & 0.005       & 0.015       & 0.025          & 0.04           & 0.03           & \textbf{0.135} \\ 
\ \ Limited             & 0.085       & \textbf{0.28} & 0.19           & 0.065          & 0.275         & 0.23  \\ 
\ \ Helpful             & \textbf{0.86} & 0.615       & 0.67           & 0.52           & 0.585          & 0.365 \\ 
\bottomrule
\end{tabular}

}
\label{table_human_eval}
\end{table*}

\end{document}